\documentclass[lettersize,journal]{IEEEtran}
\usepackage{amsmath,amsfonts}
\usepackage{algorithmic}
\usepackage{algorithm}
\usepackage{array}
\usepackage[caption=false,font=normalsize,labelfont=sf,textfont=sf]{subfig}
\usepackage{textcomp}
\usepackage{stfloats}
\usepackage{url}
\usepackage{verbatim}
\usepackage{graphicx}
\usepackage{cite}
\usepackage{color}
\usepackage{times}
\usepackage{amsmath}
\usepackage{amsthm}
\usepackage{booktabs}
\usepackage{setspace}
\usepackage{amssymb}
\usepackage{booktabs}
\usepackage{float}

\usepackage{multirow}
\usepackage{array}
\usepackage[table]{xcolor}
\begin{document}

\title{
A Unified Graph Selective Prompt Learning for Graph Neural Networks

\author{Bo Jiang, Hao Wu, Ziyan Zhang, Beibei Wang and Jin Tang}

\thanks{Bo Jiang, Hao Wu, Ziyan Zhang, Beibei Wang and Jin Tang are with the School of Computer Science and Technology, Anhui University, Hefei 230009, China (e-mail: jiangbo@ahu.edu.cn)}}

\markboth{Journal of \LaTeX\ Class Files,~Vol.~14, No.~8, August~2021}%
{Shell \MakeLowercase{\textit{et al.}}: A Sample Article Using IEEEtran.cls for IEEE Journals}

\IEEEpubid{}

\maketitle

\begin{abstract}
In recent years, 
graph prompt learning/tuning has garnered increasing attention in adapting pre-trained
models for graph representation learning. 
As a kind of universal graph prompt learning method, 
Graph Prompt Feature
(GPF) has achieved  remarkable success in adapting pre-trained
models for Graph Neural Networks (GNNs). 
By fixing 
the parameters of a pre-trained GNN model, the aim of GPF is to modify the input graph data by adding some (learnable) prompt vectors into graph node features to better align with the downstream tasks on the smaller dataset. 
However, 
existing GPFs generally suffer from two main limitations. 
First, GPFs generally focus on node prompt learning which ignore the prompting for graph edges. 
Second, existing GPFs generally conduct the prompt learning on all nodes equally which
fails to capture the importances of different nodes and may perform sensitively w.r.t noisy nodes in aligning with the downstream tasks. 
To address these issues, 
in this paper, we propose a new unified Graph Selective Prompt Feature learning (GSPF) for GNN fine-tuning. 
The proposed GSPF integrates 
the prompt learning on both graph node and edge together, which thus provides a unified prompt model for the graph data.
Moreover, 
it conducts prompt learning selectively on nodes and edges by concentrating on the important nodes and edges for prompting which thus make our model be more reliable and compact.  
Experimental results on many benchmark datasets demonstrate the effectiveness and advantages of the proposed GSPF method. 

\end{abstract}

\begin{IEEEkeywords}
Graph neural networks, Graph prompt learning, Graph pre-training, Graph fine-tuning
\end{IEEEkeywords}

\section{Introduction}
Graph Neural Networks (GNNs) are gaining popularity due to their ability to address the complex graph structures in graph data~\cite{kipf2017gcn,velickovic2018gat,0:xu2018GIN,hamilton2017graphsage}. 
However, in many applications, task-specific labels are scarce for GNN training\cite{zitnik2018prioritizing}. Inspired by the success of the `pre-training + fine-tuning' approach in Natural Language Processing (NLP) and Computer Vision (CV) fields,  
fine-tuning techniques have been applied into GNNs which provide an important solution for label scarcity issue\cite{0:hu2020strategies,hu2020Gpt-gnn,lu2021pretraingnn,0:you2020GCL,rong2020grover}. 
For example,
GraphCL\cite{0:you2020GCL} first performs unsupervised pre-training via graph data augmentation and then fine-tunes the pre-trained model on the downstream tasks.
GROVER\cite{rong2020grover} uses a Transformer based pre-training model and then fine-tunes it for molecular property prediction task. 
However,
it is known that there exist two main issues for `pre-training + fine-tuning': (i) how to reduce the misalignment gap between  the pre-training objective and the downstream task, and (ii) how to maintain the model's generalization ability in the case of small-scale downstream data~\cite{0:fang2024GPF,li2024adaptergnn}.
Recently,  in CV field\cite{Prefix-Tuning,P-Tuningv2,0:jia2022VPT,min2023recent},
prompt learning offers a promising solution to overcome the above issues, which can reduce the gap between pre-training objective and fine-tuning task by reformulating the downstream tasks to better align with the pre-trained models.
%
Inspired by this, some recent studies also attempt to design prompt learning for graph data learning~\cite{sun2023graphpromptsurvey}.
Overall, existing graph prompt learning methods can  be categorized into two types, i.e., task aligning and graph data prompting.
For task aligning prompt, some works focus on constructing the prompt templates to unify pre-training objective and downstream task. For example, GPPT~\cite{0:sun2022gppt} converts the downstream node classification task into the pre-trained link prediction task. GraphPrompt\cite{0:liu2023graphprompt} transforms both link prediction of pre-training and the downstream node classification and graph classification tasks of fine-tuning into the general subgraph similarity computation. In work~\cite{0:sun2023all}, it translates both node and edge level downstream tasks into the unified graph-level tasks. 
For graph data prompting, some works aim to incorporate prompts directly into the input data. For example, All-in-One\cite{0:sun2023all} transforms the input node features into prompt features by using prompt tokens, token structure and  inserting patterns. HetGPT\cite{0:ma2023hetgpt} introduces a small number of trainable parameters into the node features of the heterogeneous graph to generate prompted node feature vectors. GPF\cite{0:fang2024GPF} adapts pre-trained GNNs for the downstream task by applying a common prompt to all nodes' features. 
Moreover, GPF-plus\cite{0:fang2024GPF} further adjusts node features by assigning different prompts for different nodes.
In addition, GraphGLOW\cite{zhao2023graphglow} trains a shared structure learner to generate adaptive structures for the downstream target graphs. AAGOD\cite{guo2023AAGOD} proposes edge-level prompt by adjusting edge weights in the adjacency matrix with a learnable amplifier. 

However, the above existing graph prompting methods generally focus on either node prompt or edge prompt which ignores the simultaneous prompting for both of them. 
Besides, existing methods generally conduct prompting on all nodes/edges equally which fails to capture the importances/confidences of different nodes/edges and thus may perform sensitively w.r.t noisy nodes and edges.  
To address these issues, 
this paper proposes a new unified Graph Selective Prompt Feature learning (GSPF) for GNN fine-tuning. 
Specifically, GSPF constructs the graph prompt by integrating the prompting on both nodes and edges together. 
Thus, it provides a unified prompting for GNN's fine-tuning. 
Moreover, GSPF conducts prompt learning selectively on graph nodes and edges by concentrating on the important nodes and edges which thus performs more robustly and compactly w.r.t noisy and redundant nodes and edges. 

Overall, the main contributions of this paper are summarized as follows:
\begin{itemize}

  \item 
  We  propose a unified graph prompt feature learning for GNNs by conducting prompt learning on both graph nodes and edges simultaneously. 
  
  \item 
  We propose a novel graph selective prompt for graph data by considering the importances of graph nodes and edges which thus make the prompt learning be conducted more robustly and compactly.

  \item Experimental results on many benchmark datasets demonstrate the effectiveness and advantages of the proposed GSPF method under different pre-training models. 
\end{itemize}

The rest of this paper is organized  as follows. 
In Section II, we introduce some related works. In Section III, we present our GSPF method for fine-tuning GNNs. In section IV, we evaluate the effectiveness of the proposed method on several standard benchmarks. 

\section{Related Works}
\subsection{Pre-training on Graphs}
Graph pre-training methods aim to pre-train a general model and gain some general knowledge for the downstream tasks. Recent studies have proposed many representative graph pre-training strategies.
For example,  
Attribute Masking\cite{0:hu2020strategies} improves model understanding by randomly masking node or edge attributes and training the model to recover them. 
Context Prediction\cite{0:hu2020strategies} enhances the similarity between the representations of the same node's neighborhood and context graph, making that structurally similar nodes have close embeddings.
Edge Prediction\cite{hamilton2017graphsage,kipf2016gae} self-supervises the training of GNNs by predicting whether an edge exists between pairs of nodes.
Deep Graph Infomax\cite{0:velickovic2018infomax} maximizes local mutual information to obtain node embeddings that reflect global structural properties.
Graph Contrastive Learning\cite{0:you2020GCL} employs various graph augmentations and uses the contrastive loss to train models to distinguish the augmented graph views and negative examples.
GRACE\cite{Zhu:2020GRACE} generates two different perturbations of the same graph and trains models to match node representations across different views. 
GCC\cite{qiu2020GCC} introduces a contrastive learning framework that operates at the subgraph level, training models to recognize similar subgraph structures across different graphs. 

\subsection{Prompt Learning on Graphs}

 \textbf{Design a Unified Task.} Some works focus on unifying pre-training and downstream tasks through the design of pre-training frameworks and special GNN architectures\cite{0:liu2023graphprompt,0:sun2023all}. 
 For instance, 
 GPPT\cite{0:sun2022gppt} uses link prediction for pre-training and converts nodes into token pairs to transform node classification into the link prediction. 
GraphPrompt\cite{0:liu2023graphprompt} unifies pre-training and downstream tasks into a template based on computing sub-graph similarity.
 All-in-One\cite{0:sun2023all} transforms node-level and edge-level tasks into graph classification problems by constructing induced graphs.
 PRODIGY\cite{0:huang2024prodigy} proposes a prompt graph to unify the representation of node-level, edge-level and graph-level tasks in graph machine learning.
 OFA~\cite{liu2023oneforall} uses Nodes-of-Interest sub-graphs and prompt nodes to standardize different types of graph tasks into the same binary classification tasks on class nodes. 

 \textbf{Prompt on Graph Nodes.} Inspired by prompt learning in CV field\cite{Prefix-Tuning,P-tuning,bahng2022VisualPrompt,huang2023DAMVP}, some studies aim to incorporate prompts directly into the graph nodes. 
 For example, 
 All-in-One\cite{0:sun2023all} transforms the input node features into prompt features via prompt tokens, token structure and inserting patterns. DeepGPT\cite{shirkavand2023deepgpt} introduces prompt learning into the graph transformer by adding learnable prompt tokens to the input graph in each transformer layer. 
 MolCPT\cite{diao2022molcpt} presents continuous motif prompt function to augment the molecular graph with meaningful motifs in the continuous space.  
 HGPROMPT\cite{0:yu2023hgprompt} introduces feature prompts and heterogeneity prompts to modify the input node features in the subgraph readout operation and the aggregation weights of heterogeneous subgraphs. 
 GPF and GPF-plus\cite{0:fang2024GPF} propose a universal prompt learning method by introducing additional learnable parameters into the input node features.
 SUPT\cite{lee2024supt} proposes to add some prompt vectors to the input node features at the subgraph level.

 \textbf{Prompt on Graph Edges.} In addition to graph nodes, some studies also propose graph edge prompt by learning an optimized graph structure or adjusting the edge weights. 
For example, GraphGLOW\cite{zhao2023graphglow} learns an optimized graph structure via a shared structure learner. 
 AAGOD\cite{guo2023AAGOD} conducts edge-level prompts by adjusting edge weights in graph structure with a learnable amplifier for Out-of-Distribution (OOD) detection task.
 SAP\cite{ge2023sap} adds weighted connection edges between prototype nodes and original nodes in the input graph as learnable parameters. 

\begin{figure*}
	\centering
	\includegraphics[width=1\textwidth]{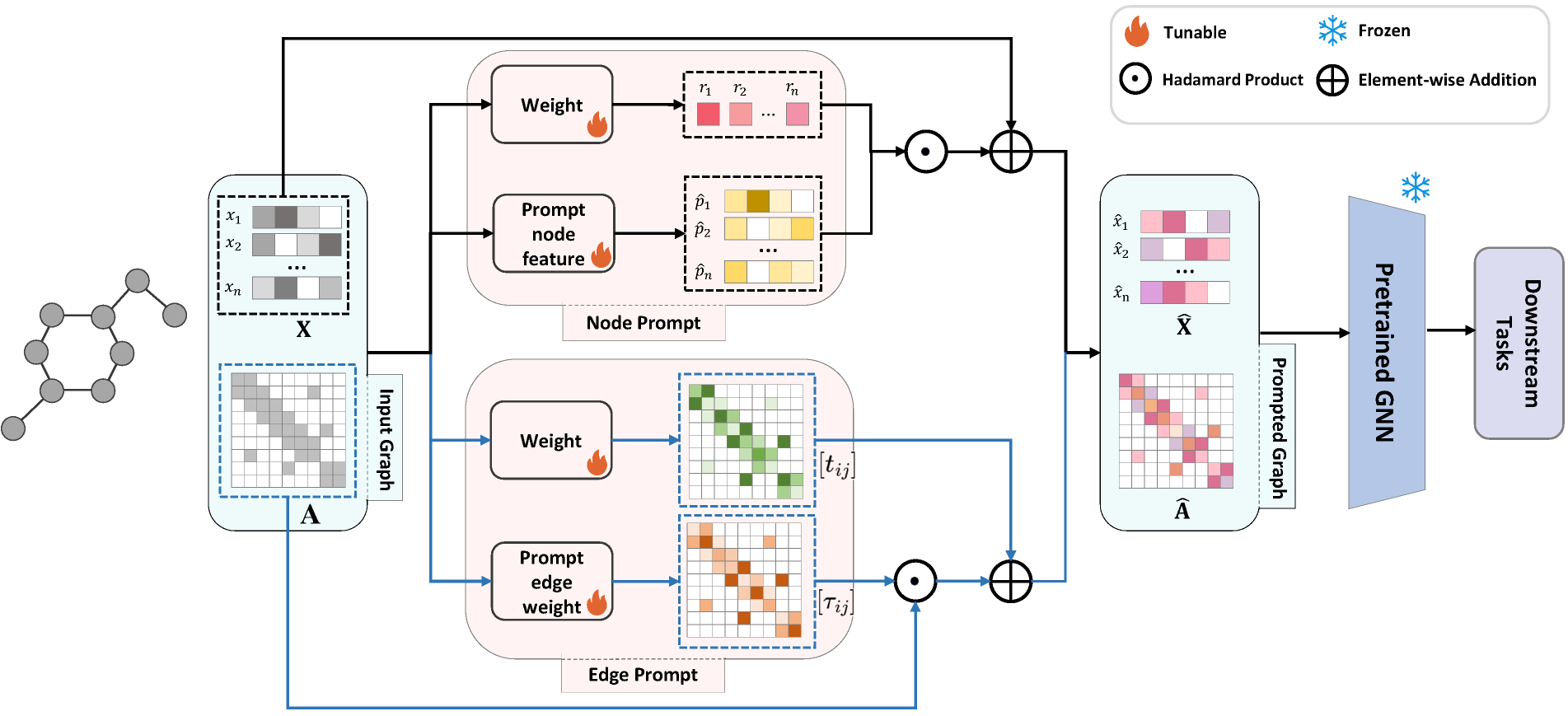}
	\caption{Illustration of our proposed  GSPF approach. It integrates  node prompt for node feature enrichment and edge prompt for adjacency matrix modification. 
 }\label{fig:frame}
\end{figure*}

\section{The Proposed Method}

\subsection{Preliminaries}

\textbf{Graph Neural Networks.} Let $G(\mathbf{A}, \mathbf{X})$ be the input graph with node set $V = \{v_1, v_2, \dots, v_n\}$ and edge set $E=\{(v_i, v_j) \mid v_i, v_j \in V\}$. The adjacency matrix $\mathbf{A}=[a_{ij}] \in \{0, 1\}^{n \times n}$ represents all edge connections between nodes, where $ a_{ij}=1$ if an edge exists between nodes $v_i$ and $v_j$, and vice versa. The node feature matrix $ \mathbf{X} \in \mathbb{R}^{n \times d}$ contains all feature vectors of nodes. Additionally, an edge feature vector $\mathbf{e}_{ij}$ is also associated with each edge $(v_i, v_j) \in E$.
Then, GNNs learn node representation by recursively aggregating the representations of its neighbors. Formally, the layer-wise message passing~\cite{gilmer2017mpnn,0:hu2020strategies} in GNNs is defined as follows, 
\begin{align}\label{EQ:}
\mathbf{x}_i^{(k)}=\mathrm{AGG}\big(\{\mathbf{x}_j^{(k-1)},\mathbf{e}_{ij}\mid v_j\in\mathcal{N}(v_i)\cup v_i\},\Theta^{(k)}\big)
\end{align}
where ${\mathbf{x}}_i^{(0)} = \mathbf{x}_i$ and ${\mathbf{x}}_i^{(k)}$ is the representation of node $v_i$ at the $k$-th layer. $\mathcal{N}(v_i)$ is the set of neighbors of $v_i$ and $\Theta^{(k)}$ denotes the learnable layer-wise weight matrix. 


\textbf{Graph Prompt Learning.} Given a pre-trained model ${f}_{\Theta}$, graph prompt learning is designed to make the pre-trained model better fit the downstream task by introducing a prompt function. Specifically, given an input graph $G (\mathbf{A}, \mathbf{X})$, one can design a modified graph $\hat{G}$ by using a task-specific graph prompting function as, 
\begin{align}\label{EQ:}
\hat{G} (\hat{\mathbf{A}}, \hat{\mathbf{X}}) = \psi_{task}\big(G(\mathbf{A}, \mathbf{X})\big)
\end{align}
The graph template $\hat{G}$ contains some learnable parameters with tunable edge weights  $\hat{\mathbf{A}}$ and node features $\hat{\mathbf{X}}$.
%
To address the specific downstream task, we fix the parameters in the pre-trained model ${f}_{\Theta}$ and only optimize the learnable prompt parameters in $\hat{G} (\hat{\mathbf{A}},\hat{\mathbf{X}})$ and the task projection head parameters $\theta$. The optimization can be described by optimizing the following objective as, 
\begin{align}\label{EQ:}
\max_{\hat{\mathbf{A}}, \hat{\mathbf{X}}, \theta} P_{f_{\Theta},\theta}(y|\hat{G})
\end{align}
where $\hat{\mathbf{A}} $ and $\hat{\mathbf{X}}$ denote the optimal adjacency matrix and feature matrix of the prompted graph $\hat{G}$. 
 

\subsection{Graph Selective Prompt Learning}
In this section, we propose our unified Graph Selective Prompt Feature learning (GSPF) to integrate both node prompt and edge prompt together for graph data, which enables the model  to adaptively adjust both the node feature and edge weight space for  downstream tasks.

\subsubsection{Selective Node Prompt}
As we know that, node prompt aims to modify node features by adding a prompt vector to each node. Given an initial node feature $\mathbf{x}_i\in\mathbb{R}^{d}$ for node $v_i$, 
the prompted feature $\hat{\mathbf{x}}_i$ can be  obtained by adding a node prompt vector $\mathbf{p}_i$ onto $\mathbf{x}_i$ as follows:
\begin{align}\label{EQ:}
\hat{\mathbf{x}}_i = \mathbf{x}_i + {\mathbf{p}}_i\,, \,\,\, i=1 \cdots n
\end{align}
where $\{\mathbf{p}_1, \mathbf{p}_2\cdots\mathbf{p}_n, \mathbf{p}_i\in \mathbb{R}^{d}\}$ denotes the collection of learnable prompt vectors. 
To reduce parameters and storage resources, one alternative way is to set $k$ prompt basis vectors $\{\mathbf{p}_1, \mathbf{p}_2\cdots\mathbf{p}_k\}
$ and then use a cross-attention mechanism to assign these prompts to all nodes~\cite{0:fang2024GPF}, i.e., we compute the prompt for each node $v_i$ as follows:
\begin{align}\label{EQ:}
&\hat{\mathbf{p}}_i = \sum_{j=1}^{k}  \frac{\exp(\mathbf{a}_j^T \mathbf{x}_i)}{\sum_{l=1}^{k} \exp(\mathbf{a}_l^T \mathbf{x}_i)} \mathbf{p}_j \\
&\hat{\mathbf{x}}_i = \mathbf{x}_i + \hat{\mathbf{p}}_i 
\end{align}
where $\{\mathbf{a}_1, \mathbf{a}_2\cdots \mathbf{a}_k \}$ are learnable parameters  in the fine-tuning process. 
However, the above Eqs.(5,6) conduct the prompt learning on all nodes equally which obviously fails to capture the weights/confidences of nodes and may perform sensitively w.r.t noisy nodes in aligning with the downstream tasks.
%
Therefore, to further refine the node prompting mechanism, we introduce an additional learnable prompt weight $r_i$ to adjust the influence of the selected prompt $\hat{\mathbf{p}}_i$ based on the node's importance. 
Specifically, $r_i$ can be learned by using a simple multi-layer perceptron (MLP) based on the node's feature $\mathbf{x}_i$ to capture the importance of each node as follows, 
\begin{align}\label{EQ:}
r_i = \sigma\big(f_{\mathrm{mlp}}(\mathbf{x}_i,{\Omega})\big)
\end{align}
where the $\Omega$ denotes the learnable parameters which are shared across all nodes and $\sigma$ denotes the Sigmoid function to enforce $r_i\in [0,1]$.  
Then, instead of using Eq.(6), we propose our selective prompt feature $\hat{\mathbf{p}}_i$ for node $v_i$ as, 
\begin{align}\label{EQ:}
\hat{\mathbf{x}}_i = \mathbf{x}_i + r_i \cdot \hat{\mathbf{p}}_i
\end{align}
where $r_i\in [0,1]$ denotes the confidence of node $v_i$ to be prompted. Here, $\hat{\mathbf{p}}_i$ is simply computed as follows, 
\begin{align}\label{EQ:}
\hat{\mathbf{p}}_i = \sum_{j=1}^{k}  \frac{\exp(\mathbf{p}_j^T \mathbf{x}_i)}{\sum_{l=1}^{k} \exp(\mathbf{p}_l^T \mathbf{x}_i)} \mathbf{p}_j
\end{align}
Note that, comparing with the previous cross-attention model Eq.(5) in work~\cite{0:fang2024GPF}, the proposed model Eq.(9) does not involve any learnable parameters which is more lightweight. 
Using the above Eqs.(8,9), one can achieve the prompt learning selectively on graph nodes. 



\subsubsection{Selective Edge Prompt}
For graph data, in addition to node prompt, one can also conduct prompt learning on graph edges. 
One simple way to implement edge prompt 
is to modify the graph adjacency matrix  by 
adding a prompt weight to each edge.
Moreover, by considering the importance of each edge, we can also conduct edge prompt learning selectively by incorporating the edge attention for each edge. 
To be specific, 
given an input edge weight $a_{ij}$, the proposed prompted edge weight  $\hat{a}_{ij}$ can be computed as follows, 
\begin{align}
\hat{a}_{ij} = {\tau}_{ij} \cdot a_{ij} + t_{ij}
\end{align}
where ${\tau}_{ij}$ denotes the learnable attention coefficient, reflecting the importance of the edge in the fine-tuning process. $t_{ij}$ denotes the prompt weight for each edge. 
To reduce the parameters, here we do not learn $\{\tau_{ij},t_{ij}\}$ directly, but learn them via the parameterized neural network. 
To be specific, we learn $\tau_{ij}$ and $t_{ij}$  as follows, 
\begin{align}
&\tau_{ij} = \mathrm{softmax}\big(\mathbf{\alpha}^T \big[\mathbf{x}_i || \mathbf{x}_j ||\mathbf{e}_{ij}\big]\big) \\
&t_{ij} = \mathbf{\beta}^T \big[\mathbf{x}_i \| \mathbf{x}_j \|\mathbf{e}_{ij}\big]
\end{align}
where $\alpha$ and $\beta $ are learnable weight vectors which are shared across all edges and $\|$ denotes the concatenation operation. 
By employing $[\tau_{ij}]$ and $[t_{ij}]$ as the layer-specific edge prompts, the prompted edge $\hat{\mathbf{A}}=[\hat{a}_{ij}]$  offer a refined mechanism to dynamically enhance the relational structure of the input graph. This facilitates a more reliable information aggregation for GNN layer-wise propagation in fine-tuning process. 

\begin{table*}
\centering
\caption{Test ROC-AUC (\%) performance on molecular prediction benchmarks using various pre-training strategies and tuning methods. The optimal results are indicated in bold.}\label{tab:mainresult}
\begin{tabular}{>{\centering\arraybackslash}m{5em}>{\centering\arraybackslash}m{4em}*{9}{>{\centering\arraybackslash}m{3em}}}
\hline
Pre-training Strategy        & Tuning Method & BBBP                              & Tox21                             & ToxCast                           & SIDER                             & ClinTox              &MUV                 &HIV                      & BACE                              & Avg.          \\ \hline
\multirow{4}{*}{Infomax}     & FT            & 68.48 \textcolor{gray}{$\pm$0.65} & 76.44 \textcolor{gray}{$\pm$0.19} & 65.29 \textcolor{gray}{$\pm$0.26} & 61.58 \textcolor{gray}{$\pm$0.80} & 74.42 \textcolor{gray}{$\pm$2.09} & 74.51 \textcolor{gray}{$\pm$1.85}& 77.28 \textcolor{gray}{$\pm$0.65}& 78.45 \textcolor{gray}{$\pm$0.56} & 72.06 \\
                             & GPF           & 66.10 \textcolor{gray}{$\pm$0.92} & 77.39 \textcolor{gray}{$\pm$0.58} & 65.38 \textcolor{gray}{$\pm$0.37} & 65.10 \textcolor{gray}{$\pm$0.61} & 72.82 \textcolor{gray}{$\pm$1.01} &80.55 \textcolor{gray}{$\pm$0.49} & 73.81 \textcolor{gray}{$\pm$2.52}& 81.81 \textcolor{gray}{$\pm$0.45} & 72.87 \\
                             & GPF-plus      & 66.68 \textcolor{gray}{$\pm$0.33} & 77.78 \textcolor{gray}{$\pm$0.36} & 65.26 \textcolor{gray}{$\pm$0.38} & 64.71 \textcolor{gray}{$\pm$0.92} & 75.57 \textcolor{gray}{$\pm$1.10} &71.28 \textcolor{gray}{$\pm$1.93}&73.50 \textcolor{gray}{$\pm$1.44}& 81.43 \textcolor{gray}{$\pm$0.46} & 72.03 \\
                             & GSPF          & \textbf{68.90} \textcolor{gray}{$\pm$0.45} & \textbf{79.69} \textcolor{gray}{$\pm$0.40} & \textbf{67.02} \textcolor{gray}{$\pm$0.22} & \textbf{66.37} \textcolor{gray}{$\pm$0.70} & \textbf{78.89} \textcolor{gray}{$\pm$0.98} &\textbf{80.64} \textcolor{gray}{$\pm$1.46}&\textbf{78.39} \textcolor{gray}{$\pm$0.74}& \textbf{84.09} \textcolor{gray}{$\pm$0.45} & 75.50 \\ \hline
\multirow{4}{*}{AttrMasking} & FT            & 68.89 \textcolor{gray}{$\pm$0.59} & 76.34 \textcolor{gray}{$\pm$0.30} & 64.75 \textcolor{gray}{$\pm$0.25} & 62.33 \textcolor{gray}{$\pm$0.66} & 77.70 \textcolor{gray}{$\pm$0.76} & 74.41 \textcolor{gray}{$\pm$1.87} & 76.67 \textcolor{gray}{$\pm$0.87}  & 80.39 \textcolor{gray}{$\pm$0.50} & 72.68 \\
                             & GPF           & 66.98 \textcolor{gray}{$\pm$1.82} & 78.19 \textcolor{gray}{$\pm$0.23} & 65.70 \textcolor{gray}{$\pm$0.22} & 68.22 \textcolor{gray}{$\pm$0.63} & 73.77 \textcolor{gray}{$\pm$1.97} & 74.74 \textcolor{gray}{$\pm$2.23}  & 72.32 \textcolor{gray}{$\pm$0.62}  & 83.88 \textcolor{gray}{$\pm$0.54} & 72.98 \\
                             & GPF-plus      & 67.29 \textcolor{gray}{$\pm$0.77} & 77.24 \textcolor{gray}{$\pm$0.39} & 65.57 \textcolor{gray}{$\pm$0.27} & 67.89 \textcolor{gray}{$\pm$0.56} & 76.90 \textcolor{gray}{$\pm$3.15} & 74.20 \textcolor{gray}{$\pm$1.74}  & 75.08 \textcolor{gray}{$\pm$1.98}  & \textbf{84.52} \textcolor{gray}{$\pm$0.81} & 73.59 \\
                             & GSPF          & \textbf{69.97} \textcolor{gray}{$\pm$1.24} & \textbf{79.98} \textcolor{gray}{$\pm$0.42} & \textbf{66.91} \textcolor{gray}{$\pm$0.37} & \textbf{69.04} \textcolor{gray}{$\pm$0.51} & \textbf{79.69} \textcolor{gray}{$\pm$0.51} & \textbf{81.02} \textcolor{gray}{$\pm$0.92}  & \textbf{78.24} \textcolor{gray}{$\pm$0.31}  & 84.51 \textcolor{gray}{$\pm$0.37} & 76.17 \\ \hline
\multirow{4}{*}{ContextPred} & FT            & 69.82 \textcolor{gray}{$\pm$1.01} & 75.83 \textcolor{gray}{$\pm$0.28} & 64.74 \textcolor{gray}{$\pm$0.21} & 62.56 \textcolor{gray}{$\pm$0.57} & 74.28 \textcolor{gray}{$\pm$1.45} & 75.99 \textcolor{gray}{$\pm$1.59}  &  79.24 \textcolor{gray}{$\pm$0.59} & 83.15 \textcolor{gray}{$\pm$0.67} & 73.20 \\
                             & GPF           & 66.48 \textcolor{gray}{$\pm$0.61} & 79.71 \textcolor{gray}{$\pm$0.33} & 66.43 \textcolor{gray}{$\pm$0.39} & 65.67 \textcolor{gray}{$\pm$0.73} & 74.80 \textcolor{gray}{$\pm$2.65} & 83.04 \textcolor{gray}{$\pm$0.43}  & 75.39 \textcolor{gray}{$\pm$1.34}  & 85.95 \textcolor{gray}{$\pm$0.79} & 74.68 \\
                             & GPF-plus      & 66.72 \textcolor{gray}{$\pm$0.42} & 79.53 \textcolor{gray}{$\pm$0.39} & 66.83 \textcolor{gray}{$\pm$0.25} & 66.01 \textcolor{gray}{$\pm$0.75} & 73.92 \textcolor{gray}{$\pm$2.80} &  82.82 \textcolor{gray}{$\pm$0.37} &  76.53 \textcolor{gray}{$\pm$1.19} & 86.00 \textcolor{gray}{$\pm$0.78}& 74.80 \\
                             & GSPF          & \textbf{72.11} \textcolor{gray}{$\pm$0.41} & \textbf{80.62} \textcolor{gray}{$\pm$0.33} & \textbf{68.53} \textcolor{gray}{$\pm$0.34} & \textbf{67.15} \textcolor{gray}{$\pm$0.89} & \textbf{78.35} \textcolor{gray}{$\pm$1.91} & \textbf{85.41} \textcolor{gray}{$\pm$1.78}  &\textbf{79.40} \textcolor{gray}{$\pm$0.31}  & \textbf{86.86} \textcolor{gray}{$\pm$0.42}& 77.30\\ \hline
\multirow{4}{*}{GraphCL}         & FT            & 69.93 \textcolor{gray}{$\pm$1.32} & 73.91 \textcolor{gray}{$\pm$0.67} & 62.70 \textcolor{gray}{$\pm$0.30} & 60.00 \textcolor{gray}{$\pm$0.68} & \textbf{77.64} \textcolor{gray}{$\pm$2.48} & 71.28 \textcolor{gray}{$\pm$2.82} & \textbf{78.71} \textcolor{gray}{$\pm$1.23} & 75.27 \textcolor{gray}{$\pm$0.74} & 71.18 \\
                             & GPF           & 70.16 \textcolor{gray}{$\pm$1.11} & 71.94 \textcolor{gray}{$\pm$0.92} & 61.47 \textcolor{gray}{$\pm$0.34} & 62.35 \textcolor{gray}{$\pm$0.65} & 71.13 \textcolor{gray}{$\pm$2.09} & 71.76 \textcolor{gray}{$\pm$1.71} &74.39 \textcolor{gray}{$\pm$0.91}  & 77.30 \textcolor{gray}{$\pm$1.49} & 70.06 \\
                             & GPF-plus      & 70.35 \textcolor{gray}{$\pm$1.07} & 72.83 \textcolor{gray}{$\pm$0.77} & 61.86 \textcolor{gray}{$\pm$0.33} & 61.65 \textcolor{gray}{$\pm$0.64} & 66.91 \textcolor{gray}{$\pm$3.01} & 70.59 \textcolor{gray}{$\pm$2.24}  &  75.28 \textcolor{gray}{$\pm$0.32} & 78.80 \textcolor{gray}{$\pm$1.12} & 69.78 \\
                             & GSPF          & \textbf{71.37} \textcolor{gray}{$\pm$1.49} & \textbf{74.98} \textcolor{gray}{$\pm$0.51} & \textbf{63.25} \textcolor{gray}{$\pm$0.37} & \textbf{64.14} \textcolor{gray}{$\pm$0.55} & 75.98 \textcolor{gray}{$\pm$3.94} & \textbf{72.07} \textcolor{gray}{$\pm$0.91} & 76.54 \textcolor{gray}{$\pm$0.69}  & \textbf{79.54} \textcolor{gray}{$\pm$0.92} & 72.23\\ \hline
\multirow{4}{*}{EdgePred}    & FT            & \textbf{70.61} \textcolor{gray}{$\pm$0.30} & 77.01 \textcolor{gray}{$\pm$0.29} & 64.98 \textcolor{gray}{$\pm$0.21} & 62.59 \textcolor{gray}{$\pm$0.24} & 75.65 \textcolor{gray}{$\pm$2.50} &  73.56 \textcolor{gray}{$\pm$1.15} &  77.46 \textcolor{gray}{$\pm$0.59} & 79.25 \textcolor{gray}{$\pm$1.16} & 72.64\\
                             & GPF           & 69.52 \textcolor{gray}{$\pm$0.27} & 79.64 \textcolor{gray}{$\pm$0.04} & 64.51 \textcolor{gray}{$\pm$0.34} & 68.22 \textcolor{gray}{$\pm$0.54} & 66.18 \textcolor{gray}{$\pm$3.10} &82.38 \textcolor{gray}{$\pm$0.70}  & 76.04 \textcolor{gray}{$\pm$0.89} & \textbf{82.63} \textcolor{gray}{$\pm$0.94} & 73.64\\
                             & GPF-plus      & 69.67 \textcolor{gray}{$\pm$0.49} & 79.94 \textcolor{gray}{$\pm$0.17} & 64.46 \textcolor{gray}{$\pm$0.32} & 66.77 \textcolor{gray}{$\pm$0.71} & 67.19 \textcolor{gray}{$\pm$3.09} & 81.97 \textcolor{gray}{$\pm$0.71}  &  76.34 \textcolor{gray}{$\pm$0.77} & 82.17 \textcolor{gray}{$\pm$1.75} & 73.56 \\ 
                             & GSPF          & 70.37 \textcolor{gray}{$\pm$0.18} & \textbf{80.79} \textcolor{gray}{$\pm$0.22} & \textbf{66.81} \textcolor{gray}{$\pm$0.29} & \textbf{68.35} \textcolor{gray}{$\pm$1.01} & \textbf{76.51} \textcolor{gray}{$\pm$1.59} & \textbf{84.40} \textcolor{gray}{$\pm$0.60}  &  \textbf{77.82} \textcolor{gray}{$\pm$1.14} & 81.75 \textcolor{gray}{$\pm$0.95} & 75.85 \\ \hline 
\end{tabular}
\end{table*}

\section{Experiment}
\subsection{Datasets}
As utilized in works~\cite{0:hu2020strategies,0:fang2024GPF}, we utilize two larger datasets for node-level self-supervised pre-training and graph-level multi-task supervised pre-training, respectively.
The pre-training dataset for node-level task contains 2 million unlabeled molecules sampled from the ZINC15~\cite{sterling2015zinc} dataset.
For graph-level multi-task supervised pre-training, we use a preprocessed ChEMBL~\cite{mayr2018ChEMBL,gaulton2012chembl} dataset, which contains 456K molecules and 1310 biochemical assays.
To demonstrate the performance and advantages of the proposed GSPF on graph-level task, we use eight molecular classification datasets from MoleculeNet~\cite{0:wu2018moleculenet} as the downstream datasets, namely BBBP, Tox21, ToxCast, SIDER, ClinTox, MUV, HIV and BACE. 
Each dataset comprises graphs where nodes represent atoms, edges represent chemical bonds and graph labels indicate some specific properties or activities. 









\subsection{Pre-training and Prompting Methods}
The baseline model GIN is pre-trained by using five unsupervised strategies, including Deep Graph Infomax (Infomax)~\cite{0:velickovic2018infomax}, Attribute Masking (AttrMasking)~\cite{0:hu2020strategies}, Context Prediction (ContextPred)~\cite{0:hu2020strategies}, Graph Contrastive Learning (GraphCL)~\cite{0:you2020GCL} and Edge Prediction (EdgePred)~\cite{hamilton2017graphsage}, to capture rich representative features from large graph datasets. 
To be specific, 
Infomax~\cite{0:velickovic2018infomax} enhances the mutual information across various graph regions, highlighting both global and local feature representations. 
AttrMasking~\cite{0:hu2020strategies} employs a technique of masking node or edge attributes and enforces the model to infer these obscured attributes, thereby improving the recognition performance.  
ContextPred~\cite{0:hu2020strategies} predicts whether neighborhood subgraphs and context graphs belong to the same node, aiming to make the embeddings of structurally similar nodes close to each other.
GraphCL~\cite{0:you2020GCL} uses the traditional contrastive learning framework with four designed graph data augmentations to train GNNs in an unsupervised way. 
EdgePred~\cite{hamilton2017graphsage} is a self-supervised method that predicts the presence of edges.
To demonstrate the benefits of our proposed GSPF method,
we first compare our method with  traditional fine-tuning strategy, i.e., fine-tuning all parameters of the pre-trained GNN model.
Besides, we compare our method with two graph prompt methods GPF and GPF-plus~\cite{0:fang2024GPF}.
For fairness, all methods use the same pre-trained model parameters provided by previous work~\cite{0:fang2024GPF}.
All comparison results reported in Table~\ref{tab:mainresult} are reproduced by the codes provided by the authors.

\subsection{Experimental Setup}
We select the widely used 5-layer Graph Isomorphism Network (GIN)~\cite{0:xu2018GIN} as the baseline GNN 
which has been proven to obtain superior performance in representation learning on several graph-level tasks~\cite{0:hu2020strategies,0:you2020GCL,0:xia2022simgrace}, as used in previous prompt learning work
 GPF and GPF-plus~\cite{0:fang2024GPF}.
In the phrase of prompt learning,
the learning rate is selected from \{1e-3, 5e-4, 1e-4\} and the weight decay parameter is selected from  \{1e-5, 1e-4, 1e-3\} to obtain the best performance. 
The layer number of Multi-Layer Perceptrons (MLPs) in the classification head is selected from 1 to 4.
The number of prompt vector counts is selected from \{1, 5, 10, 20\}.
The training epoch is selected from \{50, 100, 150, 200\}.
To ensure the reliability of the performance, 
we report the average results with standard deviation of five runs with different random seeds. 


\subsection{Comparison Results}
We compare our GSPF method with some other baseline methods and utilize ROC-AUC as the evaluation metric, as utilized in previous works~\cite{0:fang2024GPF,0:hu2020strategies}. The detailed comparison results are summarized in Table~\ref{tab:mainresult}. 
From it, we can observe that: 
(1) Compared to the traditional fine-tuning strategy, our proposed GSPF performs better on most of datasets, which demonstrates the effectiveness of applying prompt learning to the pre-trained GNNs to enhance their adaptability and performance on the specific downstream tasks.
(2) GSPF also outperforms GPF and GPF-plus~\cite{0:fang2024GPF} on most of datasets.
It indicates that,
compared to single node prompt approaches (such as GPF and GPF-plus),
integrating unified graph prompts for both nodes and edges can provide a more comprehensive and effective method to adapt pre-trained GNNs for the downstream tasks. 
(3)  GSPF achieves an average ROC-AUC score of 75.41\% on multiple different pre-training strategies, which demonstrates obvious improvements compared to the benchmark methods. 
Specifically, GSPF shows a relative increase of 4.23\% over the fine-tuning baseline's ROC-AUC of 72.35\%. 
Also, it demonstrates increase of 3.51\% and 3.66\% over GPF and GPF-plus, respectively. 
These observations indicate that GSPF provides a general and effective prompt model for most existing pre-trained GNNs. 
\begin{figure}
	\centering
	\includegraphics[width=0.45\textwidth]{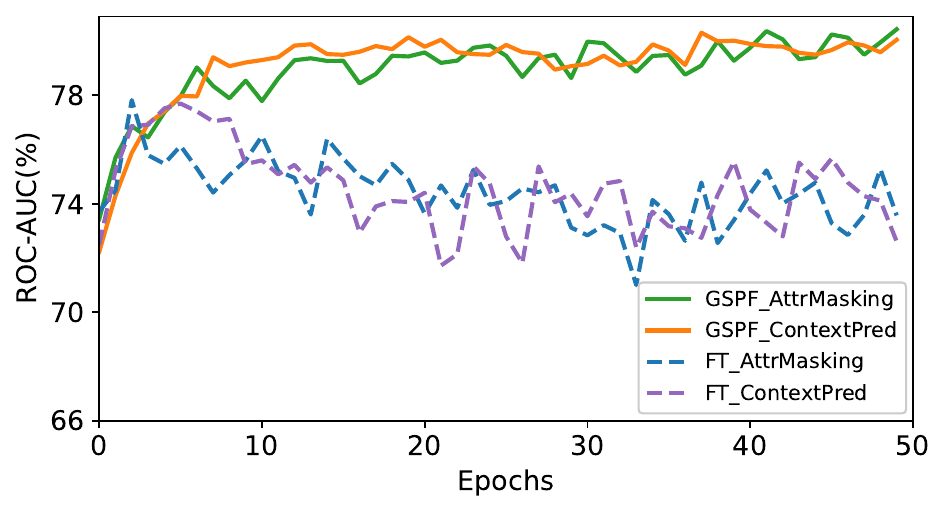}
	\caption{ROC-AUC curves of AttrMasking~\cite{0:hu2020strategies} and ContextPred~\cite{0:hu2020strategies} with FT and our GSPF method on the Tox21~\cite{0:wu2018moleculenet} dataset.}\label{fig::training_process}
\end{figure}
\begin{figure}
    \centering
    {    
        \begin{minipage}{.45\linewidth}  
        \centering
        \includegraphics[width=\linewidth]{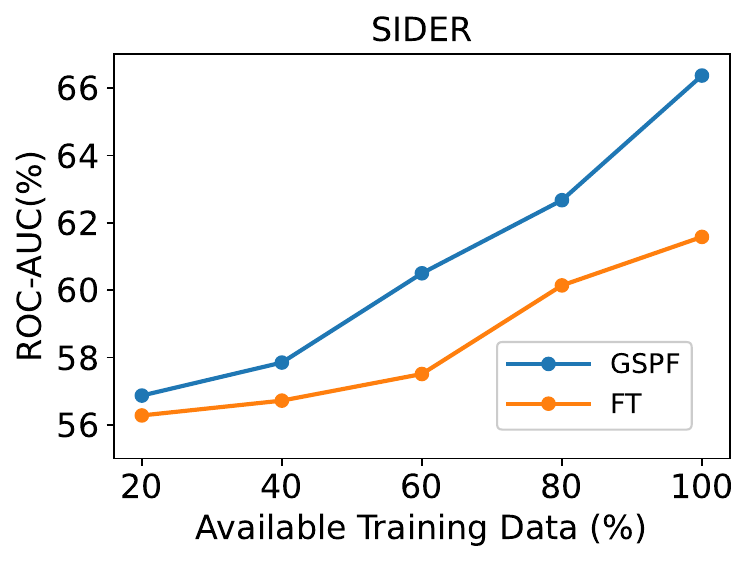}
        \end{minipage}
    }
    {    
        \begin{minipage}{.45\linewidth}  
        \centering
        \includegraphics[width=\linewidth]{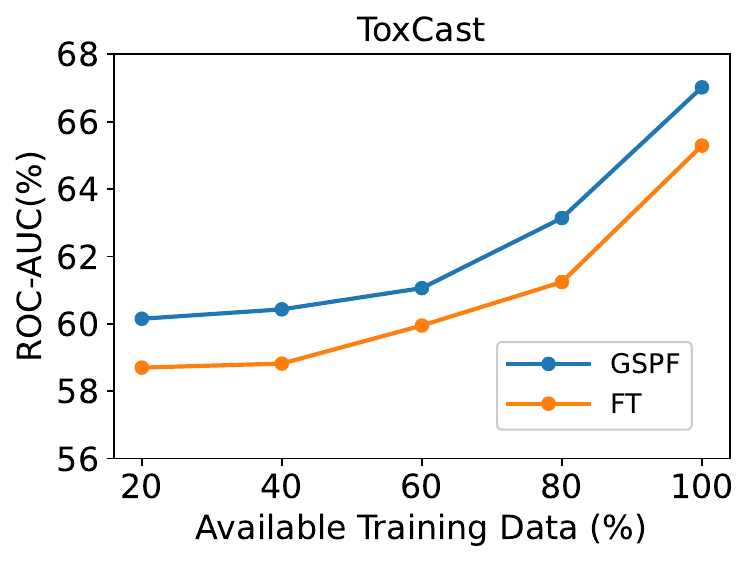}
        \end{minipage}
    }
    \caption{Comparison of test ROC-AUC (\%) performances using various training data percentages on SIDER~\cite{0:wu2018moleculenet} and ToxCast~\cite{0:wu2018moleculenet} datasets.}
    \label{fig::training_data}
\end{figure}
\begin{figure}
    \centering
    {    
        \begin{minipage}{2.7cm}  
        \centering
        \includegraphics[width=1\linewidth]{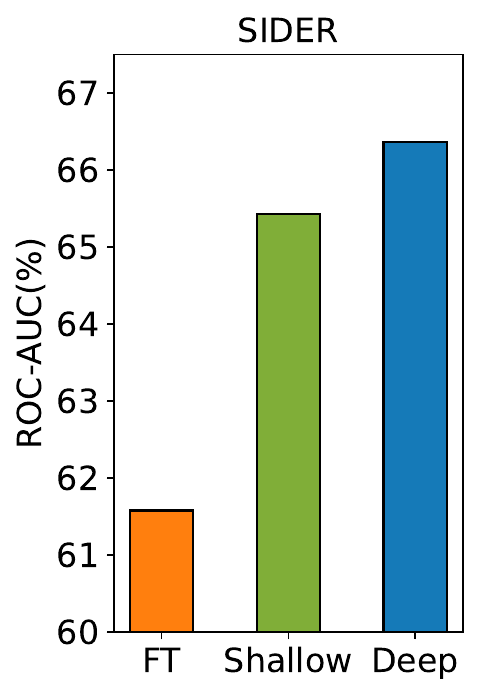}
        \end{minipage}
    }
    {    
        \begin{minipage}{2.7cm}  
        \centering
        \includegraphics[width=1\linewidth]{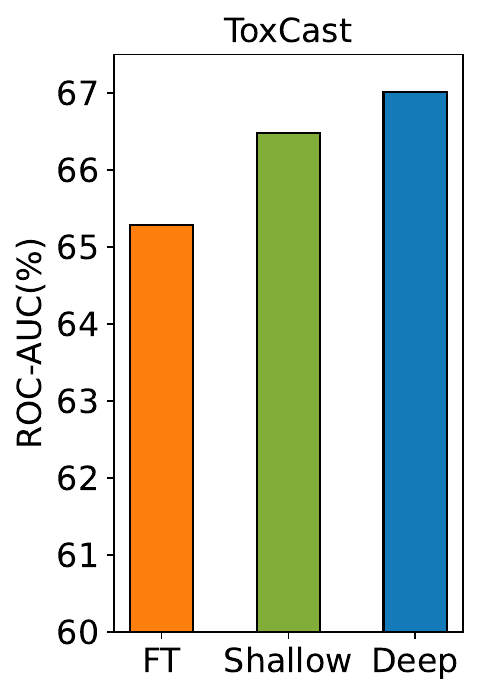}
        \end{minipage}
    }
    {    
        \begin{minipage}{2.7cm}  
        \centering
        \includegraphics[width=1\linewidth]{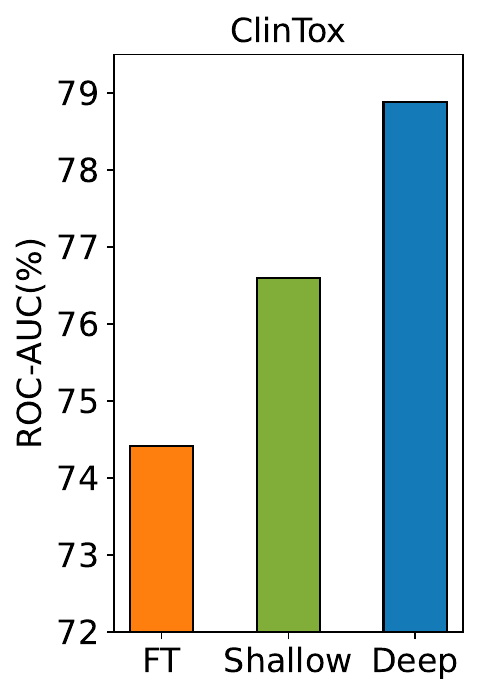}
        \end{minipage}
    }
    
    \caption{Comparative  ROC-AUC (\%) percentages for edge prompt at different layer levels: `shallow' denotes applying edge prompt only at the first GNN layer, while `deep' denotes applying edge prompt at each GNN layer.}
    \label{fig::prompt_laye_number}
\end{figure}

\begin{table}
\centering
\caption{The quantity of tunable parameters across various tuning methods.}
\label{tab:parameters}
\begin{tabular}{
>{\centering\arraybackslash}m{6em}
*{2}{>{\centering\arraybackslash}m{6em}} 
}
\hline
Tuning Method & Tunable Parameters &  Relative Ratio (\%) \\ \hline
    FT  &   $\sim$1.8M & 100      \\
    GPF & $\sim$0.3K &  0.02      \\
    GPF-plus& $\sim$3-12K &0.17-0.67  \\
    GSPF& $\sim$  9.3-15k &0.52-0.83   \\ \hline

\end{tabular}
\end{table}
\subsection{Model Analyses}
\subsubsection{Training Process Analysis}
Similar to work~\cite{0:fang2024GPF}, We compare the proposed GSPF with some other tuning strategies on the Tox21 dataset based on two pre-trained models AttrMasking~\cite{0:hu2020strategies} and ContextPred~\cite{0:hu2020strategies}. The results are 
shown in Figure~\ref{fig::training_process}.
Here, we can see that,
(1) GSPF demonstrates more stable training curves on the test set under different pre-training settings. This is mainly due to the effective integration of node and edge prompts in GSPF, which allows for better alignment and adaptation of the pre-trained models to the downstream tasks, reducing fluctuations and improving stability. 
(2) Our proposed GSPF shows the continuous improvement in ROC-AUC scores, indicating that GSPF's method of incorporating prompts at both node and edge levels can enhance the model's ability to retain the beneficial information in the training process. 


\subsubsection{Training Data Size Analysis}
To verify the generalization ability of our proposed GSPF with different amounts of training data, 
we conducted experiments on the SIDER~\cite{0:wu2018moleculenet} and ToxCast~\cite{0:wu2018moleculenet} datasets.
The results are shown in Figure~\ref{fig::training_data}.
From it, we can see that GSPF consistently outperforms traditional fine-tuning across all data sizes on both datasets. 
The possible reasons can be attributed to two aspects. 
First, compared with fine-tuning (FT), GSPF involves a smaller number of tunable parameters, which alleviates the over-fitting issue with limited training data.
Second, it also demonstrates that our proposed GSPF effectively reduces the gap between pre-training and downstream tasks, thereby achieving better performance than directly using fine-tuning method.
\begin{table*}
\centering
\caption{Ablation studies evaluate the performance of different prompts in GSPF. NP denotes the use of node prompt, EP denotes the use of edge prompt.}
\label{tab:ablation}
\begin{tabular}{
>{\centering\arraybackslash}m{2em}
>{\centering\arraybackslash}m{2em}|
*{8}{>{\centering\arraybackslash}m{4em}} 
}
\hline
NP&EP & BBBP& Tox21& ToxCast& SIDER& ClinTox& MUV& HIV& BACE \\ \hline
- & - & 69.55& 75.91& 64.49& 61.81&75.94& 73.95& 77.87& 79.30   \\
\checkmark&-& 69.79&78.37& 65.92& 66.11& 77.12& 80.01& 78.04& 83.26   \\
-&\checkmark& 69.68& 78.80& 66.25& 66.28& 76.95& 80.06& 77.98& 82.85   \\
\checkmark & \checkmark & 70.54& 79.21& 66.50& 67.01& 77.88& 80.71& 78.08& 83.35   \\
\hline

\end{tabular}
\end{table*}

\subsubsection{Edge Prompt Layer Number Analysis}
We compare the performance of applying edge prompt at each GNN layer (deep) with that of applying edge prompt only at the first GNN layer (shallow) to explore the impact of edge prompt layers on model performance. 
We conduct experiments on three datasets, including SIDER, ToxCast and ClinTox~\cite{0:wu2018moleculenet}.
We follow the same experimental setup as described previously and obtain the results in Figure~\ref{fig::prompt_laye_number}. 
From these results, we can observe the following:
(1) Both the shallow and deep versions of GSPF outperform the traditional fine-tuning method.
It demonstrates that incorporating prompts can reduce the gap between pre-training tasks and downstream tasks which results in superior performance.
(2) The deep prompting method consistently outperforms the shallow model on all datasets. 
It indicates that more edge prompt layers allow the model to capture the complex graph patterns more effectively and thus can obtain better performance.  

\subsubsection{Parameter Analysis}
We show the number of tunable parameters of various tuning methods, as summarized in Table~\ref{tab:parameters}. 
The FT method requires the full adjustment of model parameters which involves the most parameters with a parameter utilization rate of 100\%.
In contrast, the GPF method can significantly reduce the tunable parameters which utilizes only 0.02\% of the parameters required by FT. 
The tunable parameters number of GPF-plus ranges from 0.17\% to 0.67\% compared to the FT baseline.
Our proposed GSPF method integrates the prompt learning on both graph nodes and edges together which thus provides a unified prompt model for the graph data.
Moreover, GSPF conducts prompt learning selectively by concentrating on the important nodes and edges which thus makes GNN model more reliable than GPF and GPF-plus. 
Thus it requires more tunable parameters than GPF and GPF-plus.
However, it still uses less than 1\% of the FT parameters. 
This indicates that our method remains parameter-efficient when compared to FT. 

\subsection{Ablation Study}
To investigate the contributions of node prompt and edge prompt in our proposed GSPF, 
we conduct ablation studies by implementing GSPF without node prompt (NP) and edge prompt (EP) modules. The results are shown in Table~\ref{tab:ablation}.
Note that, our GSPF without both NP and EP is degenerated to the full parameter fine-tuning method (FT).
(1) The performance of our method with NP is comparable to that with EP, and both perform better than FT across all datasets. This demonstrates that the prompt strategies that concentrate on important nodes and edges are both effective. 
(2) GSPF (with both NP and EP) achieves the highest performance on all datasets. This demonstrates that our proposed GSPF effectively integrates both node and edge-level prompts which thus can effectively reduce the gap between pre-training and downstream tasks.

\section*{Conclusion}
This paper proposes 
 a novel unified Graph Selective Prompt Feature learning (GSPF) for GNN fine-tuning. 
The core aspect of the proposed  GSPF is that it integrates 
the prompt learning on both graph node and edge  together which provides a unified prompt model for the graph data.
Moreover, 
it conducts prompt learning selectively on nodes and edges by concentrating on the important nodes and edges for prompting which thus make our prompt model be more reliable and compact.  
Experimental results on many benchmark datasets demonstrate the effectiveness and advantages of the proposed GSPF method. 

\bibliography{nmfgm}
\bibliographystyle{ieeetr}

\end{document}